%% file: paper.tex
\documentclass{article}

\usepackage{microtype}
\usepackage{graphicx}
\usepackage{booktabs} 

\PassOptionsToPackage{hyphens}{url}\usepackage{hyperref}

\usepackage{xr}
\usepackage{listings}

\usepackage{adjustbox}
\usepackage{amsmath}
\usepackage{wrapfig}
\usepackage{multirow}
\usepackage{enumitem}
\setitemize{itemsep=-.5pt,topsep=-.5pt,parsep=0pt,partopsep=0pt}
\usepackage{array}
\usepackage{tikz}
\usepackage[newfloat, frozencache]{minted}
\usepackage{caption}
\usepackage{float}
\usepackage{subcaption}
\usepackage{pifont}
\usepackage{xcolor}
\usepackage{array}
\usepackage{fancyvrb}
\usepackage{fixltx2e}
\usepackage[flushleft]{threeparttable}
\newcolumntype{L}[1]{>{\raggedright\let\newline\\\arraybackslash\hspace{0pt}}m{#1}}
\newcolumntype{C}[1]{>{\centering\let\newline\\\arraybackslash\hspace{0pt}}m{#1}}
\newcolumntype{R}[1]{>{\raggedleft\let\newline\\\arraybackslash\hspace{0pt}}m{#1}}

\include{defs}

\ifminted
\else
\fi

\setlist[itemize]{noitemsep, topsep=0pt}

\usepackage[accepted]{mlsys2021}

\mlsystitlerunning{\Sys: A Compiler for Recursive Deep Learning Models}

\begin{document}

\twocolumn[
\mlsystitle{\Sys: A Compiler for Recursive Deep Learning Models}



\mlsyssetsymbol{equal}{*}

\begin{mlsysauthorlist}
\mlsysauthor{Pratik Fegade}{cmu}
\mlsysauthor{Tianqi Chen}{cmu,octoml}
\mlsysauthor{Phillip B. Gibbons}{cmu}
\mlsysauthor{Todd C. Mowry}{cmu}
\end{mlsysauthorlist}

\mlsysaffiliation{cmu}{Carnegie Mellon University, Pittsburgh, USA}
\mlsysaffiliation{octoml}{OctoML}

\mlsyscorrespondingauthor{Pratik Fegade}{ppf@cs.cmu.edu}

\mlsyskeywords{Machine Learning Inference, Recursive Models, Compilers}

\vskip 0.3in

\begin{abstract}
Optimizing deep learning models is generally performed in two steps:
(i) high-level graph optimizations such as kernel fusion and (ii) low level
kernel optimizations such as those found in vendor libraries. This
approach often leaves significant performance on the table, especially for the
case of \emph{recursive} deep learning models. In this paper, we present
\Sys, a compiler-based approach to generate highly-efficient code for
recursive models for low latency inference. Our compiler approach and
low reliance on vendor libraries enables us to perform end-to-end
optimizations, leading to up to 14X lower inference latencies over
past work, across different backends.
\end{abstract}
]



\printAffiliationsAndNotice{}  

\input{src/introduction.tex}
\input{src/overview.tex}
\input{src/recursion.tex}
\input{src/lowering_recursion.tex}
\input{src/loops.tex}
\input{src/implementation.tex}
\input{src/evaluation.tex}
\input{src/relatedwork.tex}
\input{src/conclusion.tex}
\input{src/acknowledgements.tex}

\bibliography{paper}
\bibliographystyle{mlsys2021}

\clearpage

\input{appendix.tex}

\end{document}

%% file: defs.tex
\newcommand{\Sys}{\textsc{Cortex}}

\def\zz#1{%
 \ifx\zz#1\else
   #1\linebreak[1]\expandafter\zz
 \fi}

\CustomVerbatimCommand{\ppfverb}{Verb}{commandchars=\\\{\},codes={\catcode`$=3\catcode`_=8}}

\newcommand*\circled[1]{\tikz[baseline=(char.base)]{
            \node[shape=circle,draw,thick,inner sep=1pt, scale = 0.9] (char) {#1};}}

\newcolumntype{L}[1]{>{\raggedright\let\newline\\\arraybackslash\hspace{0pt}}m{#1}}
\newcolumntype{C}[1]{>{\centering\let\newline\\\arraybackslash\hspace{0pt}}m{#1}}
\newcolumntype{R}[1]{>{\raggedleft\let\newline\\\arraybackslash\hspace{0pt}}m{#1}}
\newcommand{\gyes}[0]{\textcolor{green}{Y}}
\newcommand{\rno}[0]{\textcolor{red}{N}}
\newcommand{\ryes}[0]{\textcolor{red}{Y}}
\newcommand{\gno}[0]{\textcolor{green}{N}}
\newcommand{\opar}[0]{\textcolor{orange}{Partial}}

\newif\ifminted
\mintedtrue

\makeatletter\let\expandableinput\@@input\makeatother

%% file: src/introduction.tex
\section{Introduction}\label{sec:intro}
Deep learning models are increasingly being used in production as part
of applications such as personal assistants, self-driving
cars~\cite{sd_cars1, sd_cars2} and chatbots~\cite{cb1, cb2}. These
applications place strict requirements on the inference latency of the
models. Therefore, a wide variety of hardware substrates, including
CPUs~\cite{deepcpu}, GPUs~\cite{gpu_inf} and specialized
accelerators~\cite{tpu}, are being used in production for low latency
inference.

\begin{figure}
  \centering
  \includegraphics[width=0.75\linewidth]{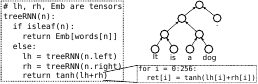}
  \vspace{-1.5mm}
  \caption{A simple recursive model. The text `It is a dog.'  is
    parsed into the parse tree which is then fed to the model.}
  \label{fig:treelstm}
\vspace{-0.05in}
\end{figure}

Reducing inference latency is especially hard for models with
recursive and other dynamic control flow. Such models have been
proposed to handle data in fields like natural language and image
processing. Textual data, represented as parse trees, can be fed to
models such as TreeLSTM~\cite{treelstm} and MV-RNN~\cite{mvrnn}.
Hierarchical and spatial relations in images can be learned by
modeling them as trees~\cite{conv_rec_dl} or
graphs~\cite{dag_rnn}. These recursive models are often extensions of
models designed for sequential data such as LSTM~\cite{lstm} and
GRU~\cite{gru}. A simple recursive model is illustrated in
Fig.~\ref{fig:treelstm}.\footnote{This is a simplified model used here
for illustrative purposes. Our evaluation is performed on actual
models.} We use this model as a running example throughout the text.

Past work on recursive and dynamic models such as DyNet~\cite{dynet1,
  dynet2}, Cavs~\cite{cavs} and PyTorch~\cite{pytorch} has relied on
hardware-specific, highly-optimized vendor libraries such as
cuDNN~\cite{cudnn} for Nvidia GPUs and MKL~\cite{mkl} for Intel
CPUs. The use of vendor libraries allows these frameworks to offer a
generic interface to users, while employing specialized and
high-performance kernel implementations in the runtime, and to
effectively utilize the wide array of backends that need to be
targeted.

Vendor libraries, however, have disadvantages in terms of \emph{model
coverage} and \emph{development effort}. As these libraries are highly
optimized, implementing them is a very intensive process. They,
therefore, contain implementations only for the most commonly used
models and kernels. For example, cuDNN contains implementations for
the LSTM and GRU models, but not for the less commonly used TreeLSTM
and MV-RNN models.

Moreover, each kernel in a vendor library is optimized \emph{in
isolation}. This often precludes optimizations such as \emph{kernel
fusion} (combining multiple kernel calls into a single call) that have
proven quite beneficial~\cite{bert_training1}.  \emph{Model
persistence} (persisting any model parameters that are reused in every
iteration of a recursive or iterative model in fast on-chip memory) is
another important optimization~\cite{deepcpu, grnn, persistent_rnns}.
But exploiting such reuse is difficult when using vendor libraries,
especially on accelerators such as GPUs with manually managed
caches~\cite{cnn_reuse, tc, tvm}. These difficulties also hold for
frameworks such as Nimble~\cite{nimble}, which relies on auto-tuned
implementations of individual kernels.

In this work, instead of relying on vendor libraries or auto-tuned kernels,
we propose a
compiler-based approach, which enables us to perform optimizations
such kernel fusion and model persistence. While there is past work
that compiles common feed forward models, applying this approach to
\emph{recursive} models has the following challenges:

\begin{enumerate}[label=\textbf{C.\arabic*}, leftmargin=2em, topsep=0pt, itemsep=-0.25em]
\item \label{ch:rec_repr} \textbf{Effective representation of
  recursive control flow:} Fig.~\ref{fig:treelstm} illustrates that
  recursive models contain dynamic control flow, along with regular
  numerical (tensor) code. Such models require an intermediate representation (IR)
  that is amenable to compiler optimizations and code generation over
  tensor computations with recursive control flow.
\item \label{ch:rec_opt} \textbf{Optimizing recursive control flow:}
  Achieving low latency inference for recursive models requires
  effective ways to execute the control flow without hindering
  optimizations such as kernel fusion.
\item \label{ch:static_opt} \textbf{Static optimizations:}
  Dynamic models are generally optimized at \emph{runtime} by
  constructing a dataflow graph that unrolls all recursion and makes
  optimizations such as \emph{dynamic batching} easier~\cite{dynet1,
    tffold}. Such optimizations have to be performed \emph{statically}
  in a compiler-based approach.
\end{enumerate}

With these challenges in mind, we present
\Sys\footnote{\textbf{CO}mpiler for \textbf{R}ecursive \textbf{T}ensor
\textbf{EX}ecution}, a compiler framework enabling users to
express iterative and recursive models and to generate efficient code
across different backends (CPUs and GPUs). To
overcome challenge~\ref{ch:rec_repr}, we observe that the control flow in
recursive models often depends solely on the input data
structure. This insight, along with a few others discussed in
\S\ref{sec:overview}, enables us to lower the recursive
computation into an efficient loop-based one (illustrated in
Fig.~\ref{fig:overview}).  To overcome~\ref{ch:rec_opt}
and~\ref{ch:static_opt}, we employ scheduling primitives to perform
optimizations such as \emph{specialization} and \emph{dynamic
batching}~\cite{dynet2, tffold}, along with compile-time optimizations
such as \emph{computation hoisting}.

\Sys's compiler-based approach enables it to optimize model
computations in an end-to-end manner, without having to treat
operators as black-box function calls, as is the case when using
vendor libraries. This enables extensive kernel fusion
(\S\ref{sec:stacked_eval}) while avoiding some overheads associated
with the dynamic batching optimization (\S\ref{sec:overall_perf}). As
part of \Sys's design, we extend a tensor compiler~\cite{halide, tvm,
  tiramisu, taco}. This enables us to reuse past work on tensor
compilers in the context of recursive models. It also opens the door
to the use of the extensive work on
auto-scheduling~\cite{halide_auto1, halide_auto2, tvm_autotune,
  tvm_autosched} for optimizing these models. Table~\ref{table:comp}
provides a qualitative comparison of \Sys~with related work on
recursive models.

\begin{table}[t]
  \vspace{-3mm}
  \caption{Comparison between \Sys~and related work on recursive models (Cavs, DyNet, Nimble
  and PyTorch).}
  \vspace{-1.5mm}
\label{table:comp}
\begin{center}
\begin{scriptsize}
\begin{tabular}{C{1cm}|C{1.4cm}C{1.5cm}C{1cm}C{1.2cm}}
\toprule
Framework & Kernel Fusion               &              Vendor Libraries & Dynamic Batching    & Model Persistence \\
\midrule
Cavs      & \opar                       & \ryes                         & \gyes            & \rno              \\
DyNet     & \rno                        & \ryes                         & \gyes            & \rno              \\
Nimble    & \opar                       & \gno                          & \rno             & \rno              \\
PyTorch   & \rno                        & \ryes                         & \rno             & \rno              \\
\midrule
\Sys      & \gyes                       & \gno                          & \gyes            & \gyes             \\
\bottomrule
\end{tabular}
\end{scriptsize}
\end{center}
\end{table}

In short, this paper makes the following contributions:
\begin{enumerate}[topsep=0pt, itemsep=-0.25em, leftmargin=1.75em]
  \item We design \Sys, a compiler-based framework that enables
    end-to-end optimization and efficient code generation for low
    latency inference of recursive deep learning models.
  \item As part of the design, we broaden the abstractions provided by
    tensor compilers and propose new scheduling primitives and
    optimizations for recursive models.
  \item We prototype the proposed framework, evaluate it against
    state-of-the-art recursive deep learning frameworks~\cite{cavs,
      dynet1,pytorch} and report significant performance gains (up to 14X) on
    Nvidia GPUs and Intel and ARM CPUs.
\end{enumerate}

%% file: src/overview.tex
\section{Overview}\label{sec:overview}
\begin{figure*}
  \centering
  \includegraphics[width=0.93\linewidth]{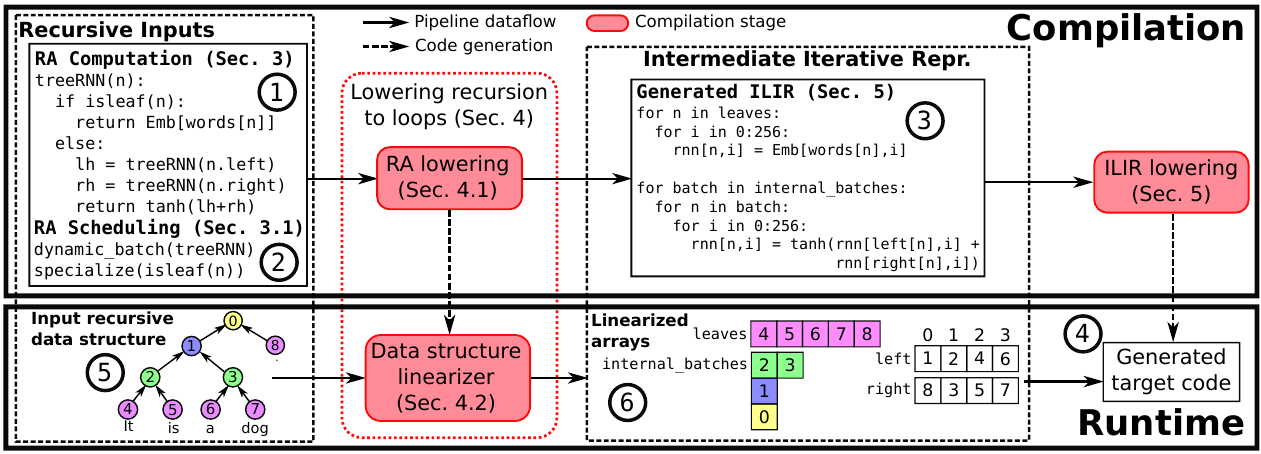}
  \vspace{-1.5mm}
  \caption{Overview of the \Sys~compilation and runtime pipeline.}
  \vspace{-8pt}
  \label{fig:overview}
\end{figure*}

Recursive deep learning models
generally traverse recursive data structures while performing tensor
computations. Efficiently executing such models is challenging because their
dynamic control flow often precludes common optimizations
such as kernel fusion. In \Sys, we observe that the control flow in
recursive models often satisfies certain properties, allowing us to
lower it to loop-based iterative control flow efficiently. In
particular, we note that a lot of recursive models have the following
properties:

\begin{enumerate}[label=\textbf{P.\arabic*}, leftmargin=2em, topsep=0pt, itemsep=-0.25em]
\item \label{prop:ds_ctrl} All control flow depends on the
  connectivity of the data structure, and not on dynamically computed
  data.
\item \label{prop:first_call} All recursive calls can be made before
  performing any tensor computation.
\item \label{prop:ind_calls} Recursive calls to the children of a data
  structure node are independent of each other: the arguments to
  one call do not depend on the results of a previous call.
\end{enumerate}

Property~\ref{prop:ds_ctrl} implies that all control flow in the model
is encapsulated in the input data
structure. Property~\ref{prop:first_call} means that computation can
start at the leaves of the data structure, moving up towards the
roots. Property~\ref{prop:ind_calls} allows us to process
sibling nodes in parallel. Taken together, these properties make it
possible to generate efficient loop-based code for these
recursive model computations.

We now look at \Sys's compilation and runtime workflows (illustrated
in Fig.~\ref{fig:overview}) that make use of these
insights. Compilation starts with the recursive model computation
\circled{1} expressed in the Recursive API (RA). The user can also
specify some scheduling primitives \circled{2} at this stage to
control how the recursive computation is lowered. The compiler then
generates Irregular Loop IR (ILIR) \circled{3} corresponding to the
input computation, according to the scheduling primitives provided by
the user. The ILIR is an extension of the IR used by tensor compilers,
designed to support additional features such as indirect memory
accesses and variable loop bounds. It is purely loop-based and data
structure agnostic. The RA lowering phase thus lowers all recursive
control flow into loops and all data structure accesses to potentially
indirect memory accesses at this stage. Loop optimizations such
unrolling, tiling, etc., as performed in tensor compilers, can be
performed here, after which target-specific code \circled{4} is
generated as part of ILIR lowering.

The runtime workflow mirrors the lowering during compilation. We start
with pointer linked recursive data structures \circled{5} such as
sequences, trees or directed acyclic graphs (DAGs), which are then
lowered to arrays \circled{6}, or in other words \emph{linearized}, by
the data structure linearizer. Such linearization makes it possible
for the generated iterative code to traverse the data structures. The
linearizer must ensure that the data dependences between the nodes of
the data structure are satisfied as it performs this lowering. Note
that the linearization stage does not involve any tensor
computations. This is because property~\ref{prop:ds_ctrl} allows us to
separate out the recursive control flow from the tensor
computation. We therefore perform linearization on the host CPU.

We now discuss each of the aforementioned compilation and execution
stages below.

%% file: src/recursion.tex
\section{Recursive API (RA)}\label{sec:recursion}
\ifminted

\begin{listing}[h]
\begin{minted}[escapeinside=||,linenos,numbersep=4pt,frame=lines,fontsize=\scriptsize]{src/ra_lexer.py:RALexer -x}
################## Model computation ##################
# H: Hidden and embedding size
# V: Vocabulary size
# N: Total number of nodes in the input data structure(s)
Tensor Emb = input_tensor((V,H))
Tensor words = input_tensor((N))

# A placeholder that represents results of recursive calls
Tensor rnn_ph = placeholder((N,H))
# Base case definition
Tensor leaf_case =
  compute((N,H), lambda n,i: Emb[words[n],i])
# Recursive body definition
Tensor lh = compute((N,H), lambda n,i: rnn_ph[n.left,i])
Tensor rh = compute((N,H), lambda n,i: rnn_ph[n.right,i])
Tensor recursive_case =
  compute((N,H), lambda n,i: tanh(lh[n, i]+rh[n, i]))
# Conditional check for the base case
Tensor body = if_then_else((N,H), lambda n,i: (isleaf(n),
                           leaf_case, recursive_case))
# Finally, create the recursion
Tensor rnn = recursion_op(rnn_ph, body)

############### RA scheduling primitives ###############
dynamic_batch(rnn)                                                |\label{line:ra_dyn}|
specialize_if_else(body)                                          |\label{line:ra_spec}|
\end{minted}
\vspace{-4mm} \captionof{listing}{Simplified implementation of the
  model in Fig.~\ref{fig:treelstm} in RA.}
\label{code:treernn_ra}
\end{listing}

\else

\begin{listing}[h]
  \begin{lstlisting}[language=Python]
################## Model computation ##################
# H: Hidden and embedding size
# V: Vocabulary size
# N: Total number of nodes in the input data structure(s)
Tensor Emb = input_tensor((V,H))
Tensor words = input_tensor((N))

# A placeholder that represents results of recursive calls
Tensor rnn_ph = placeholder((N,H))
# Base case definition
Tensor leaf_case =
  compute((N,H), lambda n,i: Emb[words[n],i])
# Recursive body definition
Tensor lh = compute((N,H), lambda n,i: rnn_ph[n.left,i])
Tensor rh = compute((N,H), lambda n,i: rnn_ph[n.right,i])
Tensor recursive_case =
  compute((N,H), lambda n,i: tanh(lh[n, i]+rh[n, i]))
# Conditional check for the base case
Tensor body = if_then_else((N,H), lambda n,i: (isleaf(n),
                           leaf_case, recursive_case))
# Finally, create the recursion
Tensor rnn = recursion_op(rnn_ph, body)

############### RA scheduling primitives ###############
dynamic_batch(rnn)
specialize_if_else(body)
\end{lstlisting}
\captionof{listing}{Model in Fig.~\ref{fig:treelstm} as expressed in
  the RA, simplified for illustration.}
\label{code:treernn_ra}
\end{listing}

\fi

\Sys~needs to have an end-to-end view of the model computation in
order to perform optimizations such as kernel fusion. Accordingly, the
input program needs to contain enough information about the tensor
operations performed in the model to enable scheduling when it is
lowered to the ILIR. Therefore, the RA models an input computation as a
DAG of operators where each operator is specified as a loop nest. This
is seen in Listing~\ref{code:treernn_ra}, which shows the simplified
model from Fig.~\ref{fig:treelstm} expressed in the RA. Along with the RA
computation, the user also needs to provide basic information about
the input data structure such as the maximum number of children per
node, and the kind of the data structure (sequence, tree or DAG). This
information is used during compilation, and can be easily verified at
runtime.

\subsection{Recursion Scheduling Primitives}\label{sec:rec_sched}
When lowering the recursive computation to loops, we need to ensure
that the data dependences between the data structure nodes are
satisfied. As these dependences generally specify only a
partial ordering on the nodes, we have significant freedom when
scheduling the computations. Different schedules may afford different
degrees of parallelism, or allow for data
reuse. Lines~\ref{line:ra_dyn} and~\ref{line:ra_spec} specify
scheduling primitives in Listing~\ref{code:treernn_ra}. We propose the
following scheduling primitives to exploit these opportunities:

\noindent\textbf{Dynamic Batching: } Dynamic batching~\cite{dynet2,
  cell_batch, tffold, cavs} involves batching operators on-the-fly to
exploit parallelism in a batch in models with dynamic control flow. As
control flow in the models we study depends only on the input data
structure (property~\ref{prop:ds_ctrl}), we perform dynamic batching
during linearization. %
With dynamic batching, nodes in a tree are processed top-to-bottom as
shown in \circled{6} in Fig.~\ref{fig:overview}.

\noindent\textbf{Specialization: } Recursive computations tend to have
frequent conditional checks to check for the base condition. These
checks can hinder optimizations such as computation hoisting and
constant propagation (\S\ref{sec:const_prop}) and have execution
overheads of their own. Thus, we allow the user to specialize the
program for the two branches of a conditional
check. Listing~\ref{code:treernn_ilir} shows the generated ILIR for
our simple recursive model. Note how it has separate loop nests for
the computation of leaves and internal nodes as the leaf check was
asked to be specialized (on line~\ref{line:ra_spec} in
Listing~\ref{code:treernn_ra}).

\begin{figure}
  \centering
  \includegraphics[width=0.80\linewidth]{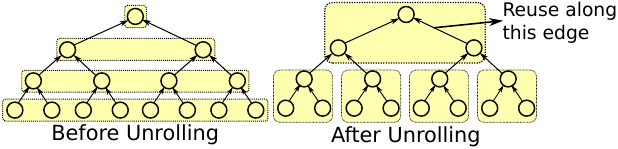}
  \vspace{-1.5mm}
  \caption{Change in execution schedule due to unrolling}
  \label{fig:unrolling}
\end{figure}

\noindent\textbf{Unrolling: } Unrolling recursion changes the order in
which nodes are processed (as illustrated in Fig~\ref{fig:unrolling}),
moving a node's computation closer in time to its children's
computation. This allows reuse of the children's hidden state via fast
on-chip caches, as opposed to the slower off-chip memory.
In Fig~\ref{fig:unrolling} (right), for example, reuse can be exploited
along every edge within a recursive call (yellow box in the figure).
Unrolling
also creates opportunities for kernel fusion as we can then fuse
operators across the children's computations.

\begin{figure}
  \centering
  \includegraphics[width=0.80\linewidth]{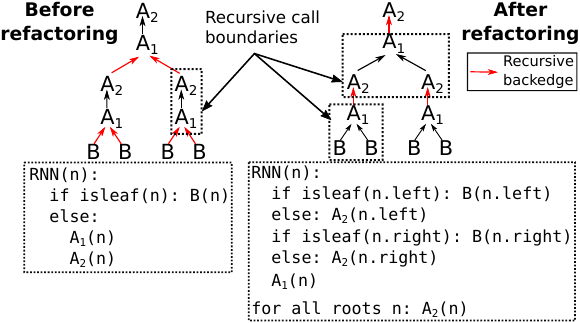}
  \vspace{-1.5mm}
  \caption{Recursive refactoring changes recursion backedge}
  \label{fig:rec_refactoring}
\end{figure}

\noindent\textbf{Recursive Refactoring: } Kernel fusion is harder to
perform across recursive call boundaries. In such cases, recursive
refactoring can be used to change the recursion backedge. Consider the
computation on the left in
Fig.~\ref{fig:rec_refactoring}. \ppfverb+A\textsubscript{1}+,
\ppfverb+A\textsubscript{2}+ and \ppfverb+B+ represent tensor
operators such that there is a dependence from
\ppfverb+A\textsubscript{1}+ to \ppfverb+A\textsubscript{2}+. In this
case, the recursive backedge goes from
\ppfverb+B+/\ppfverb+A\textsubscript{2}+ to
\ppfverb+A\textsubscript{1}+. Fusing kernels in
\ppfverb+A\textsubscript{1}(n)+ and
\ppfverb+A\textsubscript{2}(n.left)+ or
\ppfverb+A\textsubscript{2}(n.right)+ would be hard as the kernels lie
across a recursive call boundary. Refactoring changes this boundary
(the backedge now goes from \ppfverb+A\textsubscript{1}+ to
\ppfverb+A\textsubscript{2}+). Thus, \ppfverb+A\textsubscript{1}(n)+,
\ppfverb+A\textsubscript{2}(n.left)+ and
\ppfverb+A\textsubscript{2}(n.right)+ now lie in the same call and can
easily be fused.

Note that unrolling and recursive refactoring can lead to repeated and
redundant computations for DAGs as nodes can have multiple parents.
Thus, we currently support these optimizations only for trees and sequences.

%% file: src/lowering_recursion.tex
\section{Lowering Recursion to Loops}
\subsection{RA Lowering}
The lowering from the RA to the ILIR is, in essence, a lowering from
recursion to iteration. Accordingly, we make all the temporary tensors
explicit during the lowering. For instance, in the ILIR for our
running example in Listing~\ref{code:treernn_ilir}, the tensors
\verb+lh+ and \verb+rh+ are explicitly created. We also materialize
the tensor \verb+rnn+, which stores the result of the
computation. Each of the three tensors store data for each recursive
call, which in this case amounts to each tree node.


\ifminted

\begin{listing}[h]
\begin{minted}[xleftmargin=2mm, escapeinside=||,linenos,numbersep=4pt,frame=lines,fontsize=\scriptsize]{python}
for n_idx = 0:leaf_batch_size:                  |\label{line:loop1}|
  node = leaf_batch[n_idx]                      |\label{line:node_load1}|
  for i = 0:256:
    rnn[node,i] = Emb[words[node],i]

for b_idx = 0:num_internal_batches:
  for n_idx = 0:batch_sizes[b_idx]:             |\label{line:loop4}|
    node = internal_batches[b_idx,n_idx]        |\label{line:node_load2}|
    for i = 0:256:
      lh[node,i] = rnn[left[node],i]
    for i = 0:256:
      rh[node,i] = rnn[right[node],i]
    for i = 0:256:
      rnn[node,i] = tanh(lh[node,i] + rh[node,i])
\end{minted}
\vspace{-4mm}
\captionof{listing}{ILIR generated for the model in
  Fig.~\ref{fig:treelstm}}
\label{code:treernn_ilir}
\vspace{-1.5mm}
\end{listing}

\else

\begin{listing}[h]
  \begin{lstlisting}[language=Python]
for n_idx = 0:leaf_batch_size:
  node = leaf_batch[n_idx]
  for i = 0:256:
    rnn[node,i] = Emb[words[node],i]

for b_idx = 0:num_internal_batches:
  for n_idx = 0:batch_sizes[b_idx]:
    node = internal_batches[b_idx,n_idx]
    for i = 0:256:
      lh[node,i] = rnn[left[node],i]
    for i = 0:256:
      rh[node,i] = rnn[right[node],i]
    for i = 0:256:
      rnn[node,i] = tanh(lh[node,i] + rh[node,i])
\end{lstlisting}
\captionof{listing}{ILIR generated for the model in
  Fig.~\ref{fig:treelstm}}
\label{code:treernn_ilir}
\end{listing}

\fi


The scheduling primitives of recursive refactoring and unrolling are
handled by appropriately transforming the input RA computation before
the lowering. Specialized branches are handled by generating two
versions of the computation, each specialized for one target of the
branch. The data structure linearizer partitions nodes for such
specialized branches and the ILIR employs the correct version of the
computation for the respective node partition.  The lowering phase
generates the appropriate loop nest that iterates over the output of
the data structure linearizer. By default, the ILIR iterates over the
nodes, but if the user specifies dynamic batching, the ILIR iterates
over batches of nodes (as in Listing~\ref{code:treernn_ilir}).

\subsection{Data Structure Linearization}\label{sec:ds_linearization}
At runtime, the data structure linearizer traverses the input linked
structure and lays it out as arrays for the lowered loop-based
computation to iterate upon. The pseudocode for the linearizer for our
running example is shown below.

\ifminted

\begin{minted}[xleftmargin=2mm, escapeinside=||,linenos,numbersep=4pt,frame=lines,fontsize=\scriptsize]{python}
leaf_batch, internal_batches = [], [[]]
left, right = [], []

def linearizer(n):
  if isleaf(n): leaf_batch.append(node)
  else:
    linearizer(n.left)
    linearizer(n.right)
    left[n], right[n] = n.left, n.right
    internal_batches[node.height].append(node)

leaf_batch_size = len(leaf_batch)
batch_sizes = [len(b) for b in internal_batches]
num_internal_batches = len(internal_batches)
\end{minted}
\vspace{-2.8mm}

\else

\vspace{-2mm}
\begin{lstlisting}[language=Python]
leaf_batch, internal_batches = [], [[]]
left, right = [], []

def linearizer(n):
  if isleaf(n): leaf_batch.append(node)
  else:
    linearizer(n.left)
    linearizer(n.right)
    left[n], right[n] = n.left, n.right
    internal_batches[node.height].append(node)

leaf_batch_size = len(leaf_batch)
batch_sizes = [len(b) for b in internal_batches]
num_internal_batches = len(internal_batches)
\end{lstlisting}
\vspace{-2mm}

\fi

The data structure linearizer is generated during RA lowering. In the
absence of specialization and dynamic batching, the linearizer
essentially has to traverse the data structure as the input program
does, while keeping track of the order of nodes encountered. This
ordering over the nodes would satisfy data dependences and can be used
during the tensor computations. Thus, in this simple case, the data
structure linearizer is essentially the input program, stripped of all
tensor computation. For conditional checks marked for specialization,
the linearizer will separately collect nodes that follow each of the
two branches of the check. For dynamic batching, we emit code to
traverse the data structure and identify batches of nodes that can be
processed in parallel.

\subsection{Computation Hoisting and Constant Propagation}\label{sec:const_prop}
Recursive and iterative models often use an initial value for the base
case. If this initial value is same for all leaves, the same
computation is redundantly performed for all leaves. When lowering to
the ILIR, such computation is hoisted out of the recursion. We also
specially optimize the case when the initial value is the zero tensor.


%% file: src/loops.tex
\section{Irregular Loops IR (ILIR)}
We have briefly mentioned that the ILIR is an extension of the program
representation used by tensor compilers. Accordingly, computation and
optimizations are specified separately in the ILIR. The computation is
expressed as a DAG of operators, each of which produce a tensor by
consuming previously-produced or input tensors. Optimizations such as
loop tiling, loop unrolling, vectorization, etc.~can be performed with
the help of scheduling primitives.

The ILIR is generated when the recursive RA computation is lowered. As the
ILIR is loop-based and data structure agnostic, this lowering gives
rise to indirect memory accesses and loops with variable loop
bounds. Note how, in Listing~\ref{code:treernn_ilir}, the variable
\verb+node+ used to index the tensor \verb+rnn+ in the loop on
line~\ref{line:loop1} is a non-affine function of the loop variable
\verb+n_idx+. Furthermore, the loop on line~\ref{line:loop4}, which
iterates over a batch of nodes, has a variable bound, as batches can be
of different lengths. In order to support these features, we extend a
tensor compiler with (1) non-affine index expressions, (2) loops with
variable bounds, and (3) conditional operators. We describe these
modifications in further detail below.

\subsection{Indirect Memory Accesses}\label{sec:indirect_access}
We represent non-affine index expressions arising as part of indirect
memory accesses as uninterpreted functions of loop
variables~\cite{spf1}. %
Indirect memory accesses necessitate further changes, which are
described next.

\noindent\textbf{Bounds Inference: } During compilation, a tensor
compiler infers loop bounds for all operators in the input
program. For each operator $op$ producing a tensor $t$, the compiler
first computes what regions of $t$ are required for its
consumers. This quantity is then translated to the loop bounds for
$op$. In a traditional tensor compiler, this is straightforward as
there is a one-to-one correspondence between the loops of an operator
and the corresponding tensor dimensions. This is not, however, the
case with ILIR, as is apparent in
Listing~\ref{code:treernn_ilir}. Tensors \verb+lh+, \verb+rh+ and
\verb+rnn+ have two dimensions each, but the generated ILIR has three
loops for each of their corresponding operators. Therefore, we require
that the ILIR explicitly specify the relationship between tensor
dimensions and the loops in the corresponding operator's loop
nest. This is discussed further in \S\ref{sec:ap_bounds} in the
appendix.


%
\noindent\textbf{Tensor Data Layouts: } Data layouts of intermediate
tensors often need to be changed to allow for an efficient use of the
memory subsystem. %
To enable such optimizations, the ILIR exposes data layout primitives,
which allow tensor dimensions to be split, reordered and fused,
similar to the corresponding loop transformations.

\begin{figure}
  \centering
  \includegraphics[width=0.90\columnwidth]{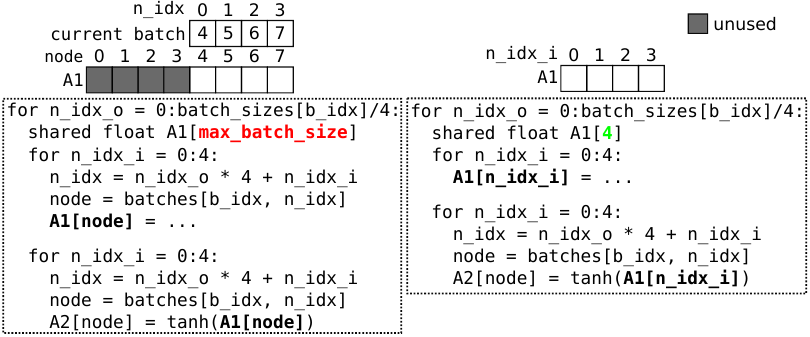}
  \vspace{-1.5mm}
  \caption{Dense indexing for intermediate tensors}
  \label{fig:layout}
\end{figure}

When an intermediate tensor is stored in a scratchpad memory, as
\verb+A1+ is Fig.~\ref{fig:layout}, indexing it with non-affine
expressions leads to a sparsely filled tensor%
. Such a sparsely filled tensor occupies excess memory, which is
problematic as scratchpad memory space is often at a premium. This is
seen on the left size of Fig.~\ref{fig:layout} where half of \verb+A1+
is unused. In such a case, we can index the tensor by the 
loop iteration space instead as seen on the right side of
Fig.~\ref{fig:layout}. Note how we now need to allocate a much smaller
tensor in the scratchpad memory. This transformation also reduces
indexing costs by turning indirect memory accesses into affine
accesses. It is exposed as a scheduling primitive as well.



\subsection{Conditional Operator}\label{sec:cond_op}
To lower conditional checks such as the \verb+isleaf+ check in our
model, we add a conditional operator to the ILIR. It takes two
sub-graphs and a conditional check as inputs and is lowered to an
\verb+if+ statement. A conditional operator would have been generated
in the ILIR for our running example if the user had \emph{not}
specialized the leaf check.

More details regarding ILIR lowering as well as a few minor
optimizations we do therein can be found in the appendix.

%% file: src/implementation.tex
\section{Implementation}\label{sec:impl}
For the purposes of evaluation, we prototype the \Sys~pipeline for the
common case. In this section, we talk about a few implementation
details regarding the same.


\noindent\textbf{RA Lowering:} As part of RA lowering, we have
implemented support for dynamic batching and specialization, for the
common case of leaf checks.

\noindent\textbf{ILIR Lowering:} We extend TVM~\cite{tvm} v0.6, a deep
learning framework and a tensor compiler. Our current prototype implementation
does not perform auto-scheduling on the generated ILIR. Therefore, the
model implementations used for evaluation were based on manually-defined
schedules. We then performed auto-tuning via grid search to
search the space of certain schedule parameters. Prior work on %
auto-scheduling is complementary to our techniques, and could readily
be applied to the prototype.


\noindent\textbf{Data Structure Linearizers:} We implemented data
structure linearizers (one each for trees and DAGs) for our
evaluation. We use a numbering scheme, described in
\S\ref{sec:ap_linearization} of the appendix, for data structures
nodes that generally reduces the costs of leaf checks and iterating
over batches.


%% file: src/evaluation.tex
\section{Evaluation}\label{sec:eval}
We now evaluate \Sys~against Cavs, DyNet and PyTorch. Cavs and DyNet
are both open source, state-of-the-art frameworks for recursive neural
networks, and have been shown to be faster than generic frameworks
like PyTorch and TensorFlow~\cite{dynet2, cavs}. PyTorch is included
for reference. We evaluate these systems on Intel and ARM CPUs and
on Nvidia GPUs.

\subsection{Experimental Setup}
\noindent\textbf{Models and Schedules: } We primarily use the models
and datasets listed in Table~\ref{table:models}. The TreeGRU model is
similar to the TreeLSTM model, except that it uses the GRU RNN
cell. The TreeLSTM and TreeGRU models were scheduled similarly to the
sequential LSTM and GRU schedules proposed in GRNN~\cite{grnn}. In the
\Sys~and PyTorch implementations for TreeLSTM, TreeGRU and DAG-RNN,
the matrix-vector multiplications involving the inputs were performed
at the beginning of the execution by a call to a matrix multiplication
kernel as in GRNN. DyNet's dynamic batching algorithm generally
performs this optimization automatically and we found that doing so
manually resulted in higher inference latencies, so we report the
automatic numbers. Unless otherwise noted, inference latencies do not
include data transfer times.

\begin{table}[t]
  \vspace{-3mm}
  \centering
  \scriptsize
  \caption{Models and datasets used in our evaluation}
  \addtolength{\tabcolsep}{-3pt}
  \begin{tabular}{L{3.2cm}L{1.2cm}L{2.95cm}}
    \toprule
    Model                                       & Short name    & Dataset used \\ \midrule
    Benchmarking model used in~\cite{tffold}    & TreeFC        & Perfect binary trees (height 7) \\ \hline
    Recursive portion of DAG-RNN~\cite{dag_rnn} & DAG-RNN       & Synthetic DAGs (size 10x10) \\ \hline
    Child-sum TreeGRU                           & TreeGRU       & Stanford sentiment treebank~\cite{sst} \\ \hline
    Child-sum TreeLSTM~\cite{treelstm}          & TreeLSTM      & Stanford sentiment treebank \\ \hline
    MV-RNN~\cite{mvrnn}                         & MV-RNN        & Stanford sentiment treebank \\
    \bottomrule
  \end{tabular}
  \addtolength{\tabcolsep}{3pt}
  \label{table:models}
\end{table}

For each model, we perform measurements for two batch sizes (1 and 10)
and two hidden sizes (256 and 512 for TreeFC, DAG-RNN, TreeGRU and
TreeLSTM and 64 and 128 for MV-RNN). The smaller and larger hidden
sizes are henceforth referred to as $h_s$ and $h_l$ respectively.

\noindent\textbf{Experimental Environment: } We use the three
environments listed in Table~\ref{table:exp_env} for the
evaluation. We use cuBLAS, Intel MKL and OpenBLAS for all BLAS needs
on the GPU, Intel and ARM backends respectively. DyNet also uses the
Eigen library. We compare against PyTorch 1.6.0, DyNet's commit
\verb+32c71acd+ (Aug.~2020) and Cavs' commit \verb+35bcc031+
(Sept.~2020).

\subsection{Overall Performance}\label{sec:overall_perf}
We compare \Sys's performance with that of PyTorch and DyNet for the
five models in Table~\ref{table:models} across the three backends.
The open-source implementation of Cavs that we evaluate against has a
few limitations---it does not fully support CPU backends, or DAG-based
models. It does not implement the lazy batching optimization as
described in the Cavs paper. It does not perform specialization nor
does it provide the user flexibility to perform the optimization
manually. In order to present a fair comparison with Cavs, we
therefore use the TreeFC, TreeGRU and TreeLSTM models on the GPU
backend, with specialization disabled in \Sys~and do not include the
input matrix-vector multiplications in both Cavs and \Sys. We were
also unable to get the streaming and fusion optimizations in Cavs
working for the TreeFC and TreeGRU models.

\begin{table}
  \vspace{-3mm}
  \centering \scriptsize
  \caption{Experimental environment}
  \begin{threeparttable}[b]
  \addtolength{\tabcolsep}{-2pt}
  \begin{tabular}{L{3.5cm}L{2.5cm}L{1.2cm}}
    \toprule
    Hardware & Software\tnote{1} & Short name \\ \midrule
    Nvidia Tesla V100 GPU (Google Cloud n1-standard-4 instance) & CUDA 10.2, cuDNN 8.0, Eigen 3.3.7 & GPU \\ \hline
    8 core, 16 thread Intel CascadeLake CPU (Google Cloud n2-standard-16 instance) & Intel MKL (v2020.0.1), Eigen (commit \verb+527210+) & Intel \\ \hline
    8 core ARM Graviton2 CPU (AWS c6g.2xlarge instance) & Eigen (commit \verb+527210+), OpenBLAS (commit \verb+5c6c2cd4+) & ARM \\
    \bottomrule
  \end{tabular}
  \addtolength{\tabcolsep}{2pt}
  \begin{tablenotes}
  \item [1] All cloud instances ran Ubuntu 18.04.
  \end{tablenotes}
  \end{threeparttable}
  \label{table:exp_env}
\end{table}

\begin{figure}
  \centering
  \includegraphics[width=0.99\columnwidth]{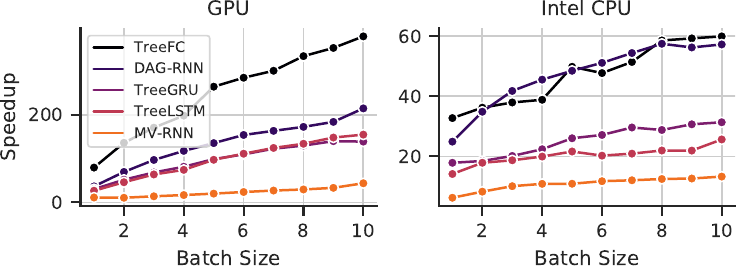}
  \vspace{-1.5mm}
  \caption{Speedup over PyTorch for hidden size $h_s$}
  \label{fig:pytorch_plot}
  \vspace{-1mm}%
\end{figure}


We first look at PyTorch. Speedups over PyTorch implementations for
the GPU and Intel backends and for hidden size $h_s$ are shown in
Fig.~\ref{fig:pytorch_plot}. PyTorch does not perform automatic
dynamic batching or kernel fusion. Due to the lack of batching, it
cannot exploit parallelism across data structure nodes leading to poor
performance. The lack of batching and kernel fusion also means that
PyTorch cannot exploit data reuse across batch elements, as well as
between multiple kernel calls. As such reuse opportunities grow with
increasing batch size, the performance gap between PyTorch and
\Sys~widens. Further, as batch sizes increase, other overheads such as
kernel invocation overheads also increase for PyTorch (as PyTorch
needs to invoke more kernels), but not for \Sys~due its extensive
kernel fusion, as we discuss later. \Sys~performs better on the GPU
backend because it can effectively utilize the higher available
parallelism on the GPU due to dynamic batching and the scratchpad
memories due to aggressive kernel fusion.

\begin{table}[t]
  \vspace{-3mm}
  \centering \scriptsize
  \caption{Cavs vs.~\Sys: Inference latencies (Cavs/\Sys) in $ms$ and
    speedups on GPU}
  \addtolength{\tabcolsep}{-3pt}
  \begin{tabular}{cc cc cc cc cc cc}
    \toprule
    \multirow{2}{6.5mm}{\centering Hidden Size} & \multirow{2}{6.5mm}{\centering Batch Size} & \multicolumn{2}{c}{TreeFC} &  \multicolumn{2}{c}{TreeGRU} &  \multicolumn{2}{c}{TreeLSTM} \\
    \cmidrule(lr){3-8}
    &  & Time & Speedup  & Time & Speedup  & Time & Speedup \\ \midrule
    \expandableinput src/cavs_table.tex
    \bottomrule
  \end{tabular}
  \addtolength{\tabcolsep}{3pt}
  \label{table:cavs_eval}
\end{table}

\begin{table*}[t]
  \vspace{-3mm}
  \centering
  \scriptsize
  \caption{DyNet vs.~\Sys: Inference latencies (DyNet/\Sys) in $ms$ and speedups across different backends}
  \begin{tabular}{ccc cc cc cc cc cc}
    \toprule
    Backend & \multirow{2}{6.5mm}{\centering Hidden Size} & \multirow{2}{6.5mm}{\centering Batch Size} & \multicolumn{2}{c}{TreeFC} &  \multicolumn{2}{c}{DAG-RNN}  & \multicolumn{2}{c}{TreeGRU}  & \multicolumn{2}{c}{TreeLSTM}  & \multicolumn{2}{c}{MV-RNN} \\
    \cmidrule(lr){4-13}
    &  &  & Time & Speedup  & Time & Speedup  & Time & Speedup  & Time & Speedup  & Time & Speedup \\ \midrule
    \input{src/dynet_table.tex}
  \end{tabular}
  \label{table:dynet_eval}
  \vspace{-2.5mm}%
\end{table*}

\begin{table*}
  \vspace{-5mm}
  \centering
  \scriptsize
  \begin{threeparttable}[b]
  \caption{Time spent ($ms$) in various activities\tnote{1} for DyNet, Cavs,
    and \Sys~for TreeLSTM on the GPU backend for batch size 10 and hidden size 256.}
  \addtolength{\tabcolsep}{0.75pt}
  \begin{tabular}{ccccccc}
    \toprule
    Framework & Dyn. batch/ Graph const. & Mem. mgmt. time (CPU/GPU) & GPU computation time & \#Kernel calls\tnote{2} & CPU CUDA API time\tnote{3} & Exe. time\tnote{4} \\
    \midrule
    DyNet & 1.21/1.82 & 1.46/1.03    & 1.71  & 389      & 12.28   & 17.381\\
    Cavs  & 0.4/-     & 0.85/1.16    & 0.71  & 122      & 9.56    & 11.57 \\
    \Sys  & 0.01/-    & -/-          & 0.32  & 1        & 0.35    & 0.35  \\
    \bottomrule
  \end{tabular}
  \addtolength{\tabcolsep}{-0.75pt}
  \begin{tablenotes}
  \item [1] The timings reported correspond to multiple runs, and were
    obtained using a combination of manual instrumentation and
    profiling using \verb+nvprof+.
  \item [2] Does not include memory copy kernels.
  \item [3] Includes all kernel calls as well as calls to
    \verb+cudaMemcpy+ and \verb+cudaMemcpyAsync+.
  \item [4] DyNet and Cavs normally execute CUDA kernels
    asynchronously. For the purposes of profiling (i.e., this table only), these calls were
    made synchronous, which leads to slower execution. Shown are
    execution times under \verb+nvprof+ profiling, provided as a
    reference.
  \end{tablenotes}
  \label{table:profile_table}
  \end{threeparttable}%
  \vspace{-2.5mm}%
\end{table*}

\begin{figure}
  \centering
  \includegraphics[width=0.99\linewidth]{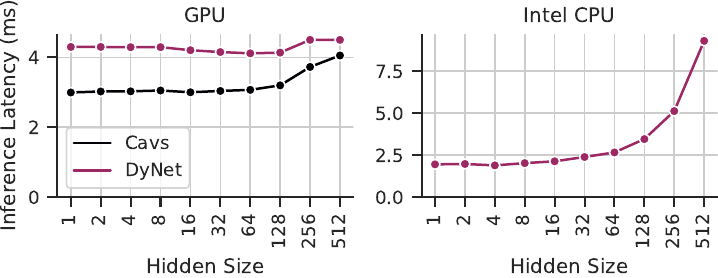}
  \vspace{-1.5mm}
  \caption{Inference latency vs.~hidden size for the recursive portion
    of TreeLSTM for batch size 10.}
  \label{fig:profile_plot}
\end{figure}

We now compare the inference latencies of \Sys~with Cavs and DyNet,
shown in Tables~\ref{table:cavs_eval} and~\ref{table:dynet_eval},
respectively. \Sys~latencies are up to 14X lower due to a number of
reasons. As compared to \Sys, Cavs and DyNet incur
significant overheads unrelated to tensor computations. This can be
seen in Fig.~\ref{fig:profile_plot}, which plots inference latency as
a function of hidden size for the TreeLSTM model\footnote{We use only
the recursive part of the TreeLSTM model, without the input
matrix-vector multiplications.} for batch size 10 for Cavs and DyNet
on the GPU and Intel backends. At low hidden sizes, the inference
latencies are quite high and are mainly comprised of overheads. As the
overheads are relatively higher for the GPU backend, we explore those
below. Apart from kernel call overheads, the discussion of the other
overheads applies to the CPU backends too.

Table~\ref{table:profile_table} lists the time spent in some runtime
components for DyNet, Cavs, and \Sys, for the same model
configuration as above on the GPU backend. DyNet and Cavs implement
generalized runtime algorithms, which cause overheads in dynamic
batching and graph construction. At runtime, DyNet constructs a
dataflow graph of tensor operators and performs dynamic batching on
the same. As compared to Cavs and \Sys, which deal with graphs
corresponding to the input data structures, DyNet therefore must
handle a much larger graph. Cavs' `think-like-a-vertex' approach also
has non-trivial overheads as compared to \Sys, which is specialized
for recursive data structures. \Sys's dynamic batching overheads are
limited to linearization, before tensor computations are executed.

As Cavs and DyNet rely on vendor libraries, they need to ensure that
inputs to batched kernel calls are contiguous in memory. The resulting
checks and memory copy operations have significant
overheads~\cite{cavs}, both on the CPU and the GPU (`Mem. mgmt. time'
in Table~\ref{table:profile_table}). As \Sys~manages the entire
compilation process, it is free from such contiguity restrictions.

\Sys~performs aggressive kernel fusion (illustrated in
Fig.~\ref{fig:fusion} and explored more
in~\S\ref{sec:ap_roofline_model} of the appendix using the roofline
model~\cite{roofline}), which has the dual effect of generating faster
GPU code (seen in the `GPU computation time' column in
Table~\ref{table:profile_table}) as well as lowering CUDA kernel call
overheads. As seen in Table~\ref{table:profile_table}, both DyNet and
Cavs execute a high number of kernel calls, which cause non-trivial
overheads as CUDA kernels calls are expensive~\cite{kernel_cost2,
  kernel_cost3}. The high number of kernel and memory copy calls also
contributes to a high amount of CPU time spent in the CUDA API as seen
in the column `CPU CUDA API time'.

\begin{figure*}
  \centering
  \includegraphics[width=1\linewidth]{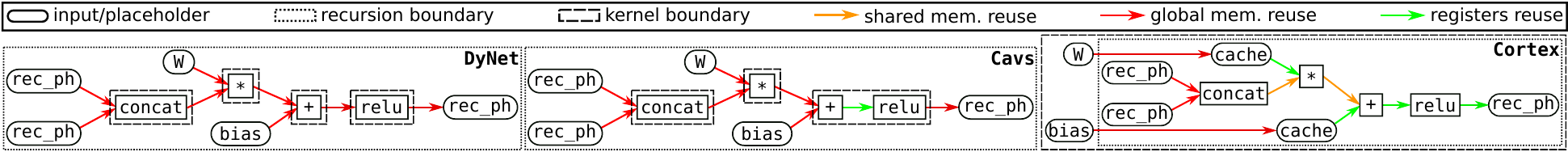}
  \vspace{-5mm}
  \caption{Kernel fusion and model persistence in \Sys: \Sys~is able
    to exploit fast on-chip memory (registers and shared memory)
    better than DyNet and Cavs. This reduces accesses to the slow
    off-chip global memory. Note also how \Sys~persists the model
    parameters (\texttt{W} and \texttt{bias}) and reuses the cached
    versions every iteration.}
  \label{fig:fusion}%
  \vspace{-2.7mm}%
\end{figure*}

\begin{figure}
  \centering
  \includegraphics[width=0.99\linewidth]{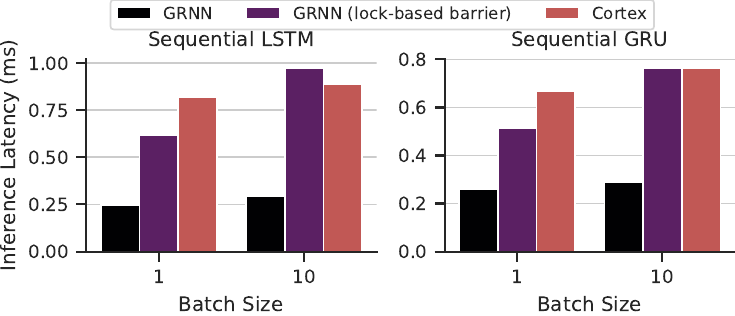}
  \vspace{-1.5mm}
  \caption{\Sys~vs.~hand-optimized GRNN code for sequence length 100
    and hidden and input sizes 256.}
  \label{fig:grnn_eval}
\end{figure}

To our knowledge, there are no hand-optimized recursive model
implementations available. Therefore, we compare \Sys~with GRNN's
hand-optimized GPU implementations of the sequential LSTM and GRU
models. These implementations use a lock-free CUDA global barrier
implementation~\cite{barrier}, which is faster than the lock-based
one~\cite{barrier} used by \Sys. For a fair comparison, we also
compare against a version of the GRNN implementations which use the
lock-based implementation. We find that \Sys-generated code performs
competitively as compared to these hand-optimized implementations
(Fig.~\ref{fig:grnn_eval}).  Notably, \Sys~can generalize these
optimizations for recursive models.

\subsection{Benefits of Optimizations}\label{sec:stacked_eval}
\begin{figure*}[h]
  \centering
  \subcaptionbox{Kernel fusion, specialization and persistence\label{fig:stacked_eval}}{
    \includegraphics[width=0.99\columnwidth]{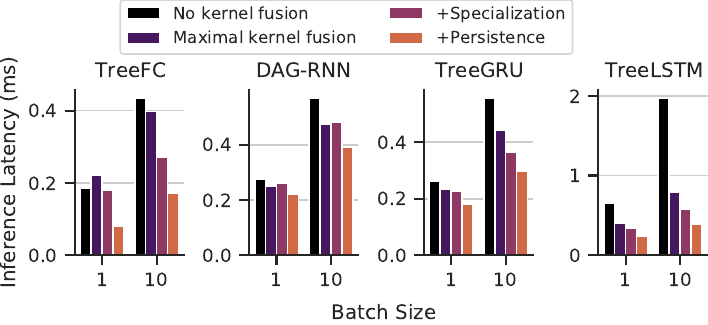}
  }%
  \hfill%
  \subcaptionbox{Unrolling\label{fig:unrolling_eval}} {
    \includegraphics[width=0.495\columnwidth]{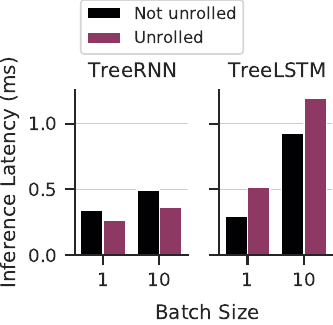}
  }%
  \hfill%
  \subcaptionbox{Recursive Refactoring\label{fig:hoisting_eval}}{
    \includegraphics[width=0.495\columnwidth]{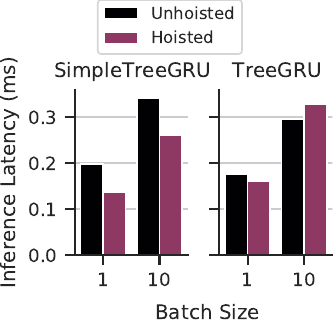}
  }%
  \vspace{-1.5mm}%
  \caption{Benefits of different optimizations on the GPU backend for
    hidden size 256.}%
  \vspace{-3mm}%
\end{figure*}

We now look at \Sys's different optimizations and their relative
benefits. Fig.~\ref{fig:stacked_eval} shows inference latencies for
different models (on GPU for hidden size 256) as we progressively
perform optimizations. Kernel fusion provides significant benefits for
all models. Fusion benefits GPUs more as GPUs have manually
managed caches, which kernels optimized in isolation cannot
exploit. Complex models such as TreeLSTM that provide more fusion
opportunities benefit more. Specialization enables computation
hoisting and constant propagation (\S\ref{sec:const_prop}), which
dramatically reduce computation in tree-based models as trees have a
larger proportion of leaves. For DAG-RNN, which performs computations
on DAGs, specialization does not lead to any speedup as
expected. Finally, model persistence leads to non-negligible
improvements by reducing accesses to the GPU global memory. We discuss
some optimization trade-offs involving register pressure in
\S\ref{sec:reg_pressure_eval} in the appendix.


\subsection{Other Scheduling Primitives}\label{sec:other_sched_eval}
We now turn to the scheduling primitives of unrolling and recursive
refactoring.

\begin{figure}
  \centering
  \includegraphics[width=0.80\linewidth]{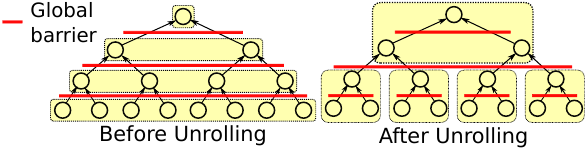}
  \vspace{-1.5mm}
  \caption{Unrolling TreeLSTM leads to additional barriers.}
  \label{fig:unroll_barriers}
\end{figure}

\noindent\textbf{Unrolling: } We evaluate unrolling on the TreeLSTM
model on the GPU backend and a hidden size of 256. In this case, after
unrolling, the cost of a barrier cannot be amortized across all nodes
in a batch, as illustrated in Fig.~\ref{fig:unroll_barriers}. This
leads to slower inference (Fig.\ref{fig:unrolling_eval}) despite the
increased data reuse and kernel fusion (\S\ref{sec:rec_sched}). We
then evaluate unrolling on the simpler TreeRNN model, which is an
extension of sequential RNNs for trees. When scheduling this model
implementation, we perform the computation for one node in one GPU
thread block, thus avoiding additional global barriers when
unrolled. Therefore, unrolling leads to a drop in the inference
latency for this model.

\noindent\textbf{Recursive Refactoring: } We evaluate recursive
refactoring on the TreeGRU model. In this case, refactoring enables us
to reduce the number of global barriers as in the GRNN GRU
implementation~\cite{grnn}. However, we find that in the case of
TreeGRU, this does not give us significant speedups
(Fig.~\ref{fig:hoisting_eval}). To explore further, we simplify the
TreeGRU model (referred to as SimpleTreeGRU\footnote{Instead of $h =
z*h_{t-1} + (1-z)*h'$, where $h'$ is the result of a linear transform,
the $h$-gate in SimpleTreeGRU is computed as $h = (1-z)*h'$.}) and
apply the same optimization again. For the case of this simplified
TreeGRU model, refactoring reduces the inference latency by about
25\%. We also use recursive refactoring in the sequential GRU model
implementation discussed above.

\subsection{Data Structure Linearization Overheads}\label{sec:linearization_eval}
The data structure linearizer (\S\ref{sec:ds_linearization}) lowers
input data structures to arrays on the host CPU, performing dynamic
batching if necessary. The table below lists linearization times (in
$\mu s$) for different models.\footnote{Models using the same dataset
are grouped together.} We find that
on the GPU backend for batch size 10 and hidden size $h_s$,
linearization overheads, as a percentage of total runtime, range from
1.2\% (for MV-RNN) to 24.4\% (for DAG-RNN). Note that the
linearization time is independent of the hidden size as no tensor
computations are performed at this stage. As \Sys~specializes for the
case of recursive data structures, the linearization overheads are
quite low.

\begin{table}[H]
  \vspace{-2mm}
  \centering
\resizebox{0.9\columnwidth}{!}{%
\begin{tabular}{cccc}
  \toprule
  \expandableinput src/linearization_table.tex
  \bottomrule
\end{tabular}
}
\vspace{-5mm}
\end{table}


\subsection{Memory Usage}
\begin{figure}
  \centering
  \includegraphics[width=0.99\linewidth]{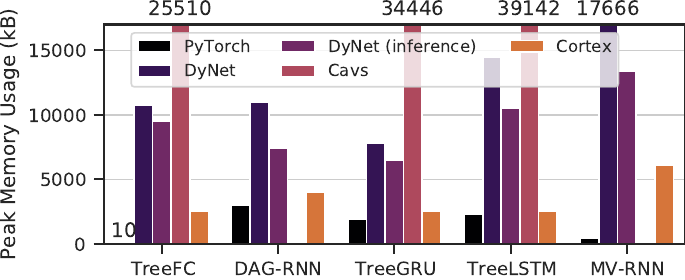}
  \vspace{-1.5mm}
  \caption{Peak GPU memory consumption in kilobytes, for batch size 10
    and hidden size $h_s$.}
  \label{fig:peak_mem}
\end{figure}

We now compare the memory consumption of \Sys~with PyTorch, DyNet and
Cavs. The peak GPU memory consumption for different models for batch
size 10 and hidden size $h_s$ is shown in
Fig.~\ref{fig:peak_mem}. PyTorch uses the least amount of memory as it
does not perform dynamic batching. DyNet and Cavs are designed for
both deep learning training and inference. As gradient computations
during training require the values of intermediate operations computed
during the forward pass, DyNet and Cavs do not free the memory used by
these intermediate tensors. Therefore, their memory consumption is
quite high as compared to \Sys, which is designed for inference. We
also compare against a version of DyNet (shown as `DyNet (inference)'
in Fig.~\ref{fig:peak_mem}) modified to simulate the deallocation of a
tensor when it is no longer needed in the forward inference
pass. Despite this deallocation, however, DyNet's memory consumption
is higher than \Sys's. \Sys~materializes fewer intermediate tensors to
the GPU's global memory due to kernel fusion
(Fig.~\ref{fig:fusion}). This reduces its memory consumption. Further,
DyNet requires extra scratch space to ensure contiguous inputs to
vendor library calls, as discussed previously.

%% file: src/dynet_table.tex
GPU & $h_s$ & 1 & 0.41/\textbf{0.08} & 5.13 & 1.79/\textbf{0.22} & 8.15 & 1.41/\textbf{0.18} & 7.69 & 1.84/\textbf{0.24} & 7.73 & 0.8/\textbf{0.34} & 2.38 \\
GPU & $h_s$ & 10 & 1.54/\textbf{0.17} & 9.26 & 3.83/\textbf{0.39} & 9.81 & 4.72/\textbf{0.35} & 13.51 & 5.28/\textbf{0.39} & 13.59 & 3.46/\textbf{0.78} & 4.42 \\
GPU & $h_l$ & 1 & 0.4/\textbf{0.12} & 3.31 & 1.78/\textbf{0.26} & 6.85 & 1.41/\textbf{0.25} & 5.66 & 1.78/\textbf{0.29} & 6.12 & 0.87/\textbf{0.39} & 2.24 \\
GPU & $h_l$ & 10 & 1.48/\textbf{0.37} & 3.97 & 3.77/\textbf{0.54} & 6.92 & 4.63/\textbf{0.75} & 6.17 & 5.1/\textbf{0.7} & 7.32 & 3.47/\textbf{1.11} & 3.14 \\ \hline
Intel & $h_s$ & 1 & 0.42/\textbf{0.12} & 3.46 & 1.12/\textbf{0.19} & 5.81 & 0.98/\textbf{0.18} & 5.42 & 1.15/\textbf{0.23} & 5.06 & 0.43/\textbf{0.29} & 1.51 \\
Intel & $h_s$ & 10 & 3.41/\textbf{0.64} & 5.29 & 6.07/\textbf{0.89} & 6.79 & 4.09/\textbf{0.89} & 4.58 & 5.59/\textbf{1.02} & 5.5 & 4.68/\textbf{1.22} & 3.83 \\
Intel & $h_l$ & 1 & 0.93/\textbf{0.42} & 2.22 & 2.21/\textbf{0.6} & 3.66 & 2.45/\textbf{0.58} & 4.19 & 2.95/\textbf{0.54} & 5.42 & 1.68/\textbf{1.08} & 1.55 \\
Intel & $h_l$ & 10 & 8.03/\textbf{2.3} & 3.49 & 11.57/\textbf{2.27} & 5.09 & 8.63/\textbf{2.97} & 2.91 & 12.36/\textbf{3.02} & 4.09 & 21.2/\textbf{7.3} & 2.9 \\ \hline
ARM & $h_s$ & 1 & 1.35/\textbf{0.21} & 6.57 & 3.48/\textbf{0.38} & 9.23 & 2.57/\textbf{0.3} & 8.49 & 2.15/\textbf{0.39} & 5.46 & 0.52/\textbf{0.4} & 1.32 \\
ARM & $h_s$ & 10 & 5.27/\textbf{1.58} & 3.32 & 11.08/\textbf{2.52} & 4.4 & 9.59/\textbf{1.81} & 5.3 & 10.59/\textbf{2.58} & 4.1 & 5.36/\textbf{2.61} & 2.05 \\
ARM & $h_l$ & 1 & 3.24/\textbf{0.79} & 4.11 & 14.39/\textbf{1.55} & 9.31 & 8.74/\textbf{0.99} & 8.8 & 6.11/\textbf{1.35} & 4.54 & 1.96/\textbf{1.95} & 1.01 \\
ARM & $h_l$ & 10 & 10.58/\textbf{6.54} & 1.62 & 26.84/\textbf{8.67} & 3.1 & 21.42/\textbf{6.08} & 3.52 & 20.11/\textbf{8.86} & 2.27 & \textbf{15.35}/16.8 & 0.91 \\ \bottomrule

%% file: src/relatedwork.tex
\section{Related Work}\label{sec:related}
\noindent\textbf{Compilers for Machine Learning: } Tensor compilers
such as TVM~\cite{tvm}, Halide~\cite{halide},
Tiramisu~\cite{tiramisu}, Tensor Comprehensions~\cite{tc} and
Taco~\cite{taco} have been well studied. There are similarities
between sparse tensor computations, as supported in Taco, and the
ILIR, which lead to similar implementation techniques. For example,
the idea of dense layouts for intermediate tensors
(\S\ref{sec:indirect_access}) is similar to the concept of workspaces
for Taco introduced in~\cite{taco_workspaces}. More generally,
however, \Sys~extends the abstractions provided by tensor compilers to
support recursive computations and develops specialized optimizations
for the same.

Deep learning compilers such as XLA~\cite{xla} and Glow~\cite{glow}
optimize static feed forward models and can perform partial kernel
fusion and code generation. Further, in~\cite{autobatch_lower}, the
authors develop techniques to efficiently lower recursion into
iterative control flow while performing dynamic batching for the XLA
toolchain. Inference engines such as TensorRT~\cite{tensorrt} and
OpenVINO~\cite{openvino} optimize model execution for inference. The
techniques we develop in this paper could be used as a low-level
backend for these deep learning compilers and
optimizers. MLIR~\cite{mlir} provides infrastructure to build deep
learning compilers, and \Sys~could potentially be built using MLIR.

\noindent\textbf{Optimizing Dynamic Neural Networks: } There is a
large body of work aimed at optimizing recursive and more generally,
dynamic neural networks.

Variants of dynamic batching have been used in frameworks such as
DyNet, Cavs, BatchMaker~\cite{cell_batch}, TensorFlow
Fold~\cite{tffold} and Matchbox~\cite{matchbox}. Unlike these,
\Sys~performs dynamic batching before any tensor computations. Model
persistence was first proposed by Persistent
RNNs~\cite{persistent_rnns}, subsequently used in GRNN~\cite{grnn} and
VPPS~\cite{vpps} and adapted for CPUs in
DeepCPU~\cite{deepcpu}. \Sys~is able to extend the
these optimizations to recursive models and formalize them as
transformation primitives in the compiler.




Nimble~\cite{nimble} adapts deep learning compiler technology for
better supporting dynamic models. Janus~\cite{janus} speculatively
creates dataflow graphs that can be optimized to accelerate dynamic
models. Similar to DyNet, this leads to overheads at
runtime. In~\cite{jeong2}, the authors extend TensorFlow's static
dataflow graph with recursion. Further, while \Sys~currently focuses
on acyclic data structures, the ILIR infrastructure could also be used
to support deep learning on more graphs, as is supported by
DGL~\cite{dgl}.


As we saw in \S\ref{sec:recursion}, \Sys~provides a lower level of
programming abstraction as compared to the frameworks mentioned
above. We believe that \Sys~could be potentially used as a backend for
these frameworks, which would alleviate the disadvantages of using
vendor libraries discussed in \S\ref{sec:intro}.





\noindent\textbf{Sparse Polyhedral Framework: } The Sparse Polyhedral
Framework (SPF)~\cite{spf1, spf2, spf3} extends the polyhedral model
for the case of sparse tensor computations. \Sys~borrows from these works
techniques
such as the use of uninterpreted functions to represent indirect
memory accesses. The data structure linearizer
in \Sys~can be viewed as an instance of the inspector-executor
technique~\cite{ins_exp}. Using this technique to lower data
structures has also been proposed in~\cite{ds_spf}.




%% file: src/conclusion.tex
\section{Conclusion}
In this paper, we presented \Sys, a compiler for optimizing recursive
deep learning models for fast inference. Eschewing vendor libraries,
\Sys's approach enables aggressive kernel fusion and end-to-end
optimizations from the recursive control flow down to the tensor
algebra computations. This allows \Sys~to achieve significantly lower
inference latencies. Past work on machine learning
compilers~\cite{relay, nimble, tf_ctrl_flow, tf_ctrl_flow_sched} as
well as on deep learning~\cite{treelstm, moe, depth_adapt_transformer}
suggests that supporting efficient execution of various kinds of
dynamism in ML models is very desirable. \Sys~demonstrates that a
fruitful way of doing this is to exploit past work on general-purpose
compilation, such as the inspector-executor technique or the sparse
polyhedral framework. We believe it is also important to expand the
scope of the highly specialized ML frameworks and techniques used
today (without compromising their ability to optimize static
feed-forward models), as we do in the case of the ILIR, for
example. In the future, we hope (i) to apply these insights to develop
similar techniques for training and serving models with potentially
non-recursive dynamic control flow and (ii) to integrate \Sys~into
higher level programming abstractions.


%% file: src/acknowledgements.tex
\section*{Acknowledgments}
This work was supported in part by grants from the National Science
Foundation and Oracle, by a
VMware University Research
Fund Award, and by the Parallel Data Lab (PDL) Consortium (Alibaba, Amazon,
Datrium, Facebook, Google, Hewlett-Packard Enterprise, Hitachi, IBM,
Intel, Microsoft, NetApp, Oracle, Salesforce, Samsung, Seagate, and
TwoSigma).
We would like to thank Chris Fallin, Dominic Chen, Hao Zhang, Graham
Neubig and Olatunji Ruwase for their suggestions and feedback on our work.

%% file: appendix.tex
\appendix
\section{ILIR Lowering}

\subsection{Uninterpreted Functions}
The ILIR extends a tensor compiler to support indirect memory accesses
and variable loop bounds. During code generation, \Sys~therefore has
to handle expressions involving such uninterpreted functions. In order
to perform simplification over such expressions, for purposes such as
proving if certain bound checks are redundant, we use the Z3 SMT
solver~\cite{z3}.

\subsection{Bounds Inference}\label{sec:ap_bounds}
\begin{figure}[h]
  \includegraphics[width=1.0\linewidth]{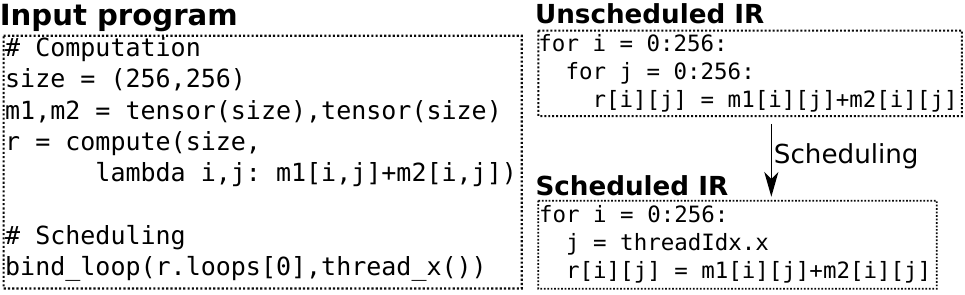}
  \caption{Element-wise matrix addition in a tensor compiler}
  \label{fig:tensor_comp_eg}
\end{figure}

We briefly mentioned in \S\ref{sec:indirect_access} how in a
traditional tensor compiler, there is a one-to-one relationship
between the dimensions of a tensor and the loops in the corresponding
operator's loop nest.\footnote{For brevity, we will not cover the case of
optimizations such as loop splitting that give rise to additional
loops. Similarly, operators involving reduction are
not covered here.} This can be seen in
Fig.~\ref{fig:tensor_comp_eg}. In the figure, two loops, each
corresponding to a dimension of the tensor \verb+r+ are generated in
IR (shown in the generated code on the right).

We also saw how in the ILIR, this relationship needs to be explicitly
specified. We do this by the way of \emph{named dimensions}. Named
dimensions are identifiers associated with tensor dimensions and
loops, which allow us to explicitly specify and keep track of
relationships between loops and tensor dimensions. Consider the ILIR
in Listing~\ref{code:ap_treernn_ilir}, which shows the named dimensions
annotated as comments. The dimensions of the tensor \verb+rnn+ are
labeled with the named dimensions \verb+d_node+ and
\verb+d_hidden+. The tensor index dimension \verb+d_node+ corresponds
to the two loop dimensions \verb+d_all_batches+ and \verb+d_batch+.

\ifminted

\begin{listing}[h]
\begin{minted}[escapeinside=||,linenos,numbersep=4pt,frame=lines,fontsize=\scriptsize]{python}
# rnn[d_node, d_hidden]
L1: for n_idx = 0:leaf_batch_size:         # d_batch
      node = leaf_batch[n_idx]
L2:   for i = 0:256:
        rnn[node,i] = Emb[words[node],i]

L3: for b_idx = 0:num_internal_batches:    # d_all_batches
L4:   for n_idx = 0:batch_sizes[b_idx]:    # d_batch
        node = internal_batches[b_idx,n_idx]
L5:     for i = 0:256:                     # d_hidden
          lh[node,i] = rnn[left[node],i]
L6:     for i = 0:256:                     # d_hidden
          rh[node,i] = rnn[right[node],i]
L7:     for i = 0:256:                     # d_hidden
          rnn[node,i] = tanh(lh[node,i] + rh[node,i])
\end{minted}
\vspace{-5mm}
\caption{ILIR generated for the model in Fig.~\ref{fig:treelstm}}
\label{code:ap_treernn_ilir}
\end{listing}

\else

\begin{listing}[h]
\begin{lstlisting}[language=Python]
# rnn[node_dim, hidden_dim]
L1: for n_idx = 0:leaf_batch_size:         # d_batch
      node = leaf_batch[n_idx]
L2:   for i = 0:256:
        rnn[node,i] = Emb[words[node],i]

L3: for b_idx = 0:num_internal_batches:    # d_all_batches
L4:   for n_idx = 0:batch_sizes[b_idx]:    # d_batch
        node = internal_batches[b_idx,n_idx]
L5:     for i = 0:256:                     # d_hidden
          lh[node,i] = rnn[left[node],i]
L6:     for i = 0:256:                     # d_hidden
          rh[node,i] = rnn[right[node],i]
L7:     for i = 0:256:                     # d_hidden
          rnn[node,i] = tanh(lh[node,i] + rh[node,i])
\end{lstlisting}
\vspace{-5mm}
\caption{ILIR generated for the model in Fig.~\ref{fig:treelstm}}
\label{code:ap_treernn_ilir}
\end{listing}

\fi

Named dimensions also make the semantic meaning of loops and index
expressions explicit. For example, the first dimension of the tensor
\verb+rnn+ is labeled \verb+d_node+ and corresponds to the space of
all nodes. It, therefore, does not make sense to index \verb+rnn+ by
\verb+b_idx+, the loop variable for the loop associated with
\verb+d_all_batches+.

\subsection{Caching Tensors Indexed by Non-Affine Expressions}\label{sec:ap_caching}
We saw in \S\ref{sec:indirect_access} how when an intermediate tensor
is stored in scratchpad memory, it can be better to index it by the
dense contiguous loop iteration space as opposed to the sparse index
space of the original tensor. A similar situation arises when
caching a tensor accessed by multiple non-affine index
expressions. Assume, for example, if we wished to cache the tensor
\verb+rnn+ in loop \verb+L4+ in Listing~\ref{code:ap_treernn_ilir},
to be used when accessing \verb+rnn[left[node],i]+ and
\verb+rnn[right[node],i]+. We create a cached tensor with an
additional dimension corresponding to the multiple non-affine index
expressions, as shown in the listing below.

\ifminted

\vspace{-2mm}
\begin{minted}[escapeinside=||,linenos,numbersep=4pt,frame=lines,fontsize=\scriptsize]{python}
  for b_idx = 0:num_internal_batches:
    for n_idx = 0:batch_sizes[b_idx]:
      node = internal_batches[b_idx,n_idx]
      for i = 0:256:
        # rnn_cache has an additional dimension
        rnn_cache[b_idx,n_idx,i,0] = rnn[left[node],i]
        rnn_cache[b_idx,n_idx,i,1] = rnn[right[node],i]

  for b_idx = 0:num_internal_batches:
    for n_idx = 0:batch_sizes[b_idx]:
      node = internal_batches[b_idx,n_idx]
      for i = 0:256:
        rnn[node,i] = tanh(rnn_cache[b_idx,n_idx,i,0] +
                           rnn_cache[b_idx,n_idx,i,1])
\end{minted}
\vspace{-2mm}

\else

\vspace{-2mm}
\begin{lstlisting}[language=Python]
  for b_idx = 0:num_internal_batches:
    for n_idx = 0:batch_sizes[b_idx]:
      node = internal_batches[b_idx,n_idx]
      for i = 0:256:
        # rnn_cache has an additional dimension
        rnn_cache[b_idx,n_idx,i,0] = rnn[left[node],i]
        rnn_cache[b_idx,n_idx,i,1] = rnn[right[node],i]

  for b_idx = 0:num_internal_batches:
    for n_idx = 0:batch_sizes[b_idx]:
      node = internal_batches[b_idx,n_idx]
      for i = 0:256:
        rnn[node,i] = tanh(rnn_cache[b_idx,n_idx,i,0] +
                           rnn_cache[b_idx,n_idx,i,1])
\end{lstlisting}
\vspace{-2mm}

\fi

\begin{figure*}
\begin{equation*}
\begin{split}
\mathcal{F} &= B \times N \times (\underbrace{4 \times H \times H}_{\text{Matrix-vector (MV) multiplication}} + \underbrace{H}_{\text{Bias computation}}) \\
\mathcal{B}_{\Sys} &= 4 \times (\underbrace{2 \times H \times H + H}_{\substack{\text{Model params: Matrix and bias} \\ \text{(read once and cached)}}} + \quad B \times N \times (\underbrace{2 \times H}_{\text{Read children hidden states}} + \underbrace{H}_{\text{Write back hidden state}})) \\
\mathcal{B}_{DyNet} &= 4 \times (\underbrace{\log_2(N) \times (2 \times H \times H + H)}_{\substack{\text{Model params: Matrix and bias} \\ \text{(read for every dyn. batch)}}} \quad + \quad B \times N \times (\underbrace{2 \times H}_{\substack{\text{Read children} \\ \text{hidden states}}} + \underbrace{H}_{\substack{\text{Write back} \\ \text{MV results}}} + \underbrace{H}_{\text{Read MV result}} + \underbrace{H}_{\substack{\text{Write back} \\ \text{hidden state}}})) \\
\mathcal{B}_{PyTorch} &= 4 \times (\underbrace{B \times N \times (2 \times H \times H + H)}_{\substack{\text{Model params: Matrix and bias} \\ \text{(read for every node)}}} \quad  + \quad B \times N \times (\underbrace{2 \times H}_{\substack{\text{Read children} \\ \text{hidden states}}} + \underbrace{H}_{\substack{\text{Write back} \\ \text{MV results}}} + \underbrace{H}_{\text{Read MV result}} + \underbrace{H}_{\substack{\text{Write back} \\ \text{hidden state}}})) \\
\end{split}
\end{equation*}
\caption{The operational intensities for PyTorch, DyNet and \Sys, for the TreeFC model. Here, $N$ is the number of nodes in a tree, $B$ is the batch size and $H$ is the hidden size.}
\vspace{-8pt}
\label{fig:roofline}
\end{figure*}

\subsection{Barrier Insertion}
We need to insert synchronization barriers and memory fences when
threads read data written by other threads. This is true on CPUs as
well as on accelerators such as GPUs. The barrier insertion pass in
TVM does well on tensor programs that do not have loop-carried
dependencies. Specifically, given a loop-carried dependence, the pass
conservatively places barriers in the innermost loop, as opposed to
placing it in the body of the loop that actually carries the
dependence. This can lead to unnecessary barriers, leading to inflated
runtimes.

As we iterate sequentially either over data structure nodes (when
dynamic batching is not performed) or batches of nodes (when dynamic
batching is performed), the data dependencies between a node and its
children manifest as loop-carried dependencies in the generated ILIR
code. This can be seen in the generated ILIR for the running example,
in Listing~\ref{code:ap_treernn_ilir}. In the listing, the data
written to tensor \verb+rnn+ in loops \verb+L2+ and \verb+L7+ is read
by loops \verb+L5+ and \verb+L6+. This dependence only exists across a
node and its children. We are also guaranteed, by the properties
described in \S\ref{sec:overview} and the way the data structure
linearizer works, that no node in a batch may be a child of any other
node in the same batch. Thus, the dependence is carried by loop
\verb+L3+, and not by loop \verb+L4+.

Given this dependence, we would need a barrier at the start of every
iteration of loop \verb+L3+. However, the conservative barrier
insertion pass in TVM instead places a barrier in the body of loop
\verb+L4+. We therefore designed a modification to the pass to insert
the barrier in the outer loop, which actually carries the dependence.

\subsection{Other Optimizations during ILIR Lowering}
Below, we discuss two minor optimizations and scheduling knobs we
implemented.

\noindent\textbf{Loop Peeling: } The generated ILIR in \Sys~involves
loops with variable loop bounds. Splitting such loops gives rise to
bounds checks in the bodies of the loops. We employ loop peeling to
ensure that such checks are only employed for the last few iterations
of the loop.

\noindent\textbf{Rational Approximations of Nonlinear Functions: } We
use rational approximations for the $tanh$ and $sigmoid$ functions,
which makes exploiting SIMD instructions on CPUs easier.

\section{Data Structure Linearization}\label{sec:ap_linearization}
In our data structure linearizers, when lowering a pointer linked data
structure to arrays, we associate the nodes with integer
identifiers. When doing so for the case of dynamic batching, we ensure
that nodes in a batch are numbered consecutively and higher than their
parents. This enables us to lower the batches into two arrays ---
\verb+batch_begin+ and \verb+batch_length+, which store the starting
node and the length, respectively, of every batch. Thus, node \verb+n+
is in batch \verb+i+ if \verb+batch_begin[i]+ \textless= \verb+n+
\textless~\verb|batch_begin[i] + batch_length[i]|.  This numbering
scheme also ensures that all leaf nodes are numbered higher than all
internal nodes. This reduces the cost of checking if a node is a
leaf. When nodes are numbered in this way, a leaf check involves a single
comparison as opposed to a memory load (to load the number of children of
a node under question, for example) and a comparison in the case where
the numbering were arbitrary. This scheme thus generally reduces the
overheads of iterating over batches and performing leaf checks.

\section{Roofline Performance Analysis for TreeFC Model}\label{sec:ap_roofline_model}

The roofline model~\cite{roofline} is a simple analytical performance
model that can be used to quantify the amount of reuse exploited by a
given computation kernel. As part of this model, the reuse exploited
by a kernel is captured in the operational intensity ($\mathcal{O}$)
of that kernel. This metric is computed as the amount of computation
performed per byte transferred from the memory. Below, we analyze the
PyTorch, DyNet and \Sys~implementations of the simple TreeFC model
using the roofline model.

Let $N$ be the number of nodes in a tree, $B$ be the batch
size and $H$ be the hidden size. In Fig.~\ref{fig:roofline}, we
compute the total number of floating point operations ($\mathcal{F}$)
in the model, which remains constant across the three frameworks, and
the total number of bytes ($\mathcal{B}$) read or written to the
off-chip memory.

Assuming, $N, H  \gg B \ge 1$ and $N \approx H = N_0$, which is the case
for our evaluation of the TreeFC model when $H = h_s$, we obtain

\begin{equation*}
\begin{split}
\mathcal{O}_{\Sys}  &= \frac{\mathcal{F}}{\mathcal{B}_{\Sys}} \approx  \frac{B \times N_0}{3 \times B + 2} \\
\mathcal{O}_{DyNet}   &= \frac{\mathcal{F}}{\mathcal{B}_{DyNet}} \approx  \frac{B \times N_0}{5 \times B + 8 \times \log_2(N_0)} \\
\mathcal{O}_{PyTorch} &= \frac{\mathcal{F}}{\mathcal{B}_{PyTorch}} \approx  0.5
\end{split}
\end{equation*}

As can be seen, $\mathcal{O}_{\Sys} > \mathcal{O}_{DyNet}
> \mathcal{O}_{PyTorch}$, suggesting that \Sys~generated kernels
exploit more data reuse as compared to DyNet and PyTorch. One should
note that this is a simple model that does not take into account other
overheads associated with DyNet and PyTorch such as the kernel call
and dynamic batching overheads discussed
in \S\ref{sec:overall_perf}.

\section{Register Pressure in CUDA}\label{sec:reg_pressure_eval}
\Sys-generated CUDA kernels are often large, due to optimizations such
as aggressive kernel fusion, loop peeling, loop unrolling and
recursive unrolling. Furthermore, model persistence uses GPU registers
to persist model weights. These factors lead to high register
pressure. %
We find that recursive unrolling precludes us from using persistence
for the TreeLSTM and TreeRNN models discussed in
\S\ref{sec:other_sched_eval}. Similarly, we note that we cannot apply
the loop peeling and model persistence optimizations in the case of
the Tree\-LSTM model at the same time. In our schedules, we have
explored this trade-off space and evaluated on the best performing
schedule. We note that techniques developed in past work such
as~\cite{reg_pressure1} and~\cite{reg_pressure2} can potentially be
applied in our context to alleviate this issue.

